\documentclass[letterpaper, 10 pt, conference]{ieeeconf} 
\IEEEoverridecommandlockouts
\usepackage{graphicx}
\usepackage{epsfig}
\usepackage{bm}
\usepackage{xcolor}
\usepackage{listings}
\usepackage[inkscapelatex=false]{svg}
\lstdefinestyle{terminal}{
    backgroundcolor=\color{gray!15},
    basicstyle=\ttfamily\small,
    frame=single,
    framerule=0pt,
    xleftmargin=0pt,
    xrightmargin=0pt,
    breaklines=true,
    showstringspaces=false,
    keepspaces=true,
    numbersep=5pt
}
\usepackage{times}
\usepackage{amsmath}
\usepackage{amssymb}
\usepackage{tabularx}
\usepackage{booktabs}
\usepackage{float}
\usepackage{algpseudocode}
\usepackage{algorithm}
\usepackage{balance}
\usepackage{hyperref}

\newcommand{\pos}{\mathbf{x}}
\newcommand{\posdot}{\dot{\mathbf{x}}}
\newcommand{\vel}{\mathbf{v}}
\newcommand{\veldot}{\dot{\mathbf{v}}}
\newcommand{\faero}{\mathbf{f}_a}
\newcommand{\taua}{\bm{\tau}_\mathrm{a}}
\newcommand{\tauext}{\bm{\tau}_\mathrm{ext}}
\newcommand{\g}{\mathbf{g}}
\newcommand{\zb}{\mathbf{z}_B}
\newcommand{\fext}{\mathbf{f}_\mathrm{ext}}

\newcommand{\w}{\mathbf{\Omega}}
\newcommand{\wdot}{\dot{\mathbf{\Omega}}}
\newcommand{\J}{\mathbf{J}}
\newcommand{\invJ}{\mathbf{J}^{-1}}
\newcommand{\kt}{k_t}
\newcommand{\ktau}{k_{\tau}}
\newcommand{\kdx}{k_{d,x}}
\newcommand{\kdy}{k_{d,y}}
\newcommand{\kdz}{k_{d,z}}
\newcommand{\kh}{k_{h}}
\newcommand{\windvel}{\bm{v}_{w}}

\newcommand{\msp}{\bm \omega}

\def\lone{{\mathcal{L}_1}}

\newcommand{\ethresh}{e_{\mathrm{th}}}

\title{A Simulation Evaluation Suite for Robust Adaptive Quadcopter Control}

\author{Dingqi Zhang$^1$, Ran Tao$^2$, Sheng Cheng$^2$, Naira Hovakimyan$^2$, and Mark W. Mueller$^1$
\thanks{$^1$The authors are with the High Performance Robotics Lab, Dept. of Mechanical Engineering, UC Berkeley. Contact at \{dingqi, mwm\}@berkeley.edu}.
\thanks{$^2$The authors are with Advanced Controls Research Laboratory, Dept. of Mechanical Engineering, University of Illinois Urbana-Champaign, Contact at \{rant3, chengs, nhovakim\}@illinois.edu}
}

\begin{document}

\maketitle

\begin{abstract}
Robust adaptive control methods are essential for maintaining quadcopter performance under external disturbances and model uncertainties. However, fragmented evaluations across tasks, simulators, and implementations hinder systematic comparison of these methods. This paper introduces an easy-to-deploy, modular simulation testbed for quadcopter control, built on \textit{RotorPy}, that enables evaluation under a wide range of disturbances such as wind, payload shifts, rotor faults, and control latency. The framework includes a library of representative adaptive and non-adaptive controllers and provides task-relevant metrics to assess tracking accuracy and robustness. The unified modular environment enables reproducible evaluation across control methods and eliminates redundant reimplementation of components such as disturbance models, trajectory generators, and analysis tools. We illustrate the testbed’s versatility through examples spanning multiple disturbance scenarios and trajectory types, including automated stress testing, to demonstrate its utility for systematic analysis. Code is available at \url{https://github.com/Dz298/AdaptiveQuadBench}.
\end{abstract}

% \begin{IEEEkeywords}
% Robust Control, Adaptive Control, Quadrotors, Benchmarking
% \end{IEEEkeywords}

%%%%%%%%%%%%%%%%%%%%%%%%%%%%%%%%%%%%%%%%%%%%%%%%%%%%%%%%%%%%%%%%%%%%%%%%%%%%%%%%

\section{Introduction}
Tracking trajectories in the presence of external disturbances and model uncertainties is a fundamental challenge in quadcopter control. Disturbances such as wind gusts, payload variations, rotor degradation, and communication latency are common in real-world settings, especially in applications like aerial logistics and infrastructure inspection. While accurate modeling can improve control performance, capturing all relevant dynamics, including time-varying forces and unmodeled aerodynamics, is often infeasible in practice. As a result, robust and adaptive control strategies have been widely explored to ensure reliable performance under uncertainty~\cite{SAVIOLO202345learningagilereview}.

A range of control approaches have been developed to address these challenges. Classical methods such as PID~\cite{johnson2005pid}, geometric control on $\mathrm{SE}(3)$~\cite{lee2011geo}, and model predictive control (MPC)~\cite{camacho2013mpc-book,rawlings2020mpc-book} offer strong performance under nominal conditions but can degrade in the presence of model mismatch. To improve robustness, techniques such as $\lone$ adaptive control~\cite{hovakimyan2010l1}, adaptive geometric control~\cite{wu2024l1quad,goodarzi2015geoa}, and adaptive incremental nonlinear dynamic inversion (INDI)~\cite{smeur2015adaptINDI} have been proposed to compensate for uncertainty in mass, inertia, aerodynamic drag, and external forces. In parallel, learning-based controllers trained in simulation using reinforcement learning or imitation learning have shown promise in transferring to real-world flight through techniques such as domain randomization~\cite{zhang2024learningbasedquadcoptercontrollerextreme,kaufmann2022benchmark} and residual learning~\cite{Bauersfeld__2021,zhang2024proxflyrobustcontrolclose}.

To validate these robust adaptive control methods, simulation plays a critical role. It enables Monte Carlo analysis across randomized scenarios, and scalable experiments beyond what is feasible on hardware. Despite significant progress, simulation-based evaluation remains fragmented.
For instance, \cite{hanover2021performance} assesses MPC augmented by $\lone$ adaptive control under model mismatch conditions using \textit{RotorS}~\cite{Furrer2016rotorS}, an open-source simulator based on Gazebo and ROS.
Similarly, \cite{sun2024comparativestudynonlinearmpc} compares nonlinear MPC with differential-flatness-based control under model uncertainties using the simulation in \textit{Agilicious}~\cite{Foehn22scienceagilicious}, a full-stack framework designed for agile flight.
While these studies provide valuable insights, the need to repeatedly develop simulation environments for similar tasks can lead to overlapping efforts.
In addition, existing simulators are frequently tied to specific application such as hardware integration (\cite{Foehn22scienceagilicious,Furrer2016rotorS}), reinforcement learning training (\cite{song2020flightmare,panerati2021learningtofly}) or agricultural inspection~\cite{zha2024agrifly}.
As a result, evaluation tools are typically embedded within broader simulation frameworks, making it challenging for researchers to decouple them from the rest of the software stack.
This separation between evaluation tools and simulation environments demands a high level of expertise, hindering accessibility for both newcomers and experienced researchers aiming for benchmarking.
\begin{figure}
    \centering
    \includegraphics[width=\columnwidth]{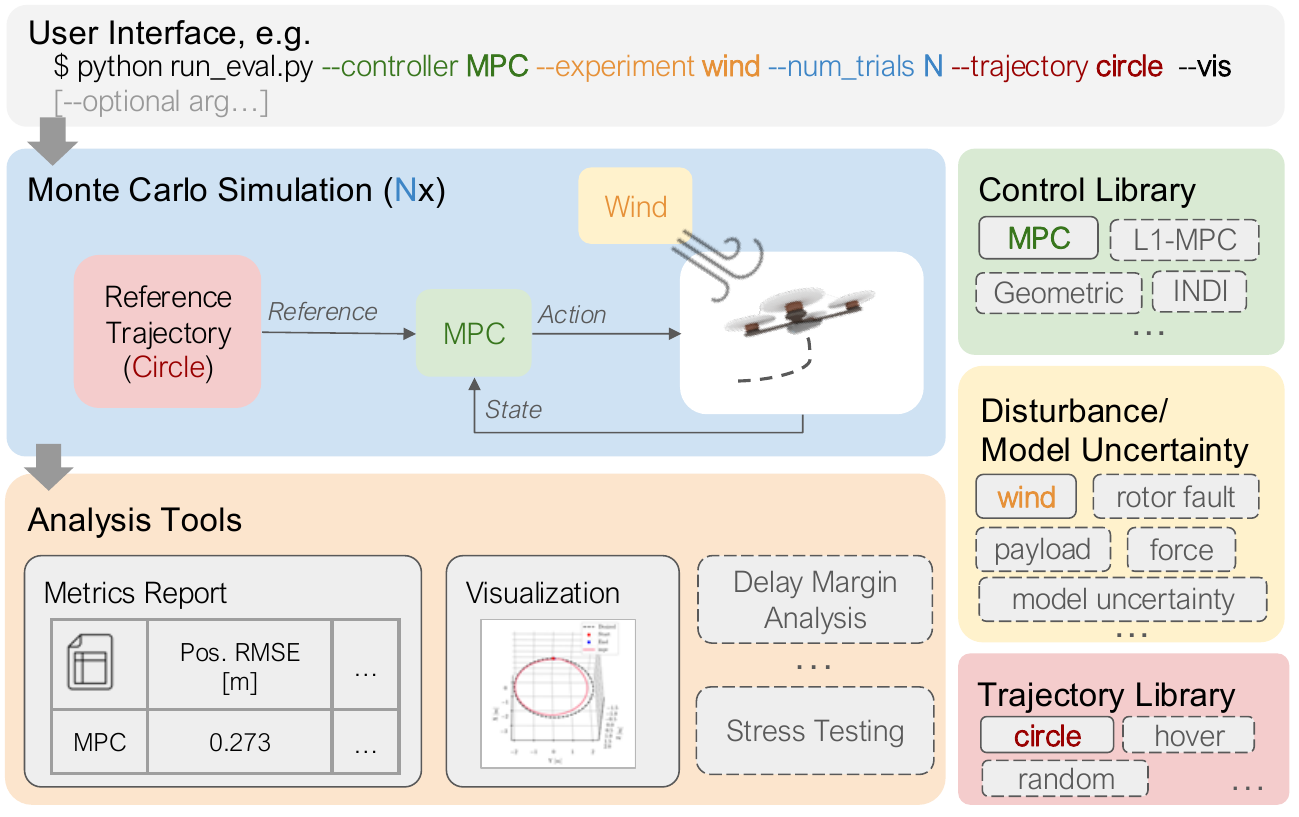}    \caption{Overview of our simulation testbed. The framework consists of a dynamics simulator based on \textit{RotorPy}~\cite{folk2023rotorpy}, a modular control interface supporting both adaptive and non-adaptive controllers, and evaluation tools for trajectory generation, disturbance injection, performance metrics, and visualization. Users interact with the system through a unified configuration interface that specifies controller selection, disturbance type, trajectory, and evaluation mode.}
    \label{fig:system-overview}
\end{figure}
The closest work in this direction is \textit{Safe Control Gym}~\cite{yuan2022safecontrolgymunifiedbenchmarksuite}, which provides a standardized evaluation environment for safe and robust control. However, its underlying dynamics are simplified (e.g., a 2D quadcopter model), limiting its ability to capture dynamics such as the aerodynamic and inertial effects crucial for realistic evaluation.

To address these challenges, we introduce a modular simulation framework that enables reproducible and systematic benchmarking of quadcopter controllers. 
Built on top of the lightweight and open-source simulator \textit{RotorPy}~\cite{folk2023rotorpy}, the testbed supports the evaluation of controllers across diverse tasks and includes a modular control library featuring representative robust adaptive control methods for quadcopters.

In summary, this paper provides: \begin{enumerate} 
% \item A modular simulation framework that enables reproducible evaluation of quadcopter control methods under diverse disturbances and model uncertainties;
% \item An open-source control library comprising adaptive controllers, enabling rapid prototyping and development for research and educational applications;
% \item A suite of task-relevant metrics and automated stress-testing tools for quantifying tracking accuracy, robustness margins, and failure thresholds.
\item A simulation benchmarking framework for quadcopter control that supports systematic, reproducible evaluation under disturbances and model uncertainties.
\item An open-source control library containing representative robust adaptive control methods for quadcopters.
\item A configurable suite of disturbance and uncertainty models, including wind, payload shifts, rotor fault, and control latency.
\item A set of evaluation tools for performance and robustness analysis.
\end{enumerate}
The framework is designed to support reproducible, structured benchmarking of quadcopter control strategies under diverse and challenging conditions, without favoring a specific controller or architecture.

\section{Framework Overview}

In this section, we detail the setup of our simulation testbed, covering quadcopter dynamics, the control pipeline, evaluation tasks, and associated performance metrics.  
Additionally, we describe the implementation of external disturbances and model uncertainties introduced into the plant.  
An overview of the system is provided in Figure~\ref{fig:system-overview}. 
The testbed is designed to support consistent evaluation of diverse control methods, enabling users to conduct their own comparisons across disturbances and uncertainties.

\subsection{Quadcopter Dynamics}\label{sec:quaddyna}
While our testbed builds on \textit{RotorPy}’s base dynamics, we summarize the model here to clarify the modifications introduced for adaptive control evaluation, following standard formulations of quadcopter flight dynamics~\cite{Mueller2025Dynamics}.
In particular, we extend \textit{RotorPy} by: (i) introducing external force and torque to the dynamics; (ii) incorporating a motor effectiveness vector to simulate rotor-level asymmetries; and (iii) adopting a planar drag formulation that has been shown effective on hardware experiments in~\cite{sun2024comparativestudynonlinearmpc, zhang2024learningbasedquadcoptercontrollerextreme}. 

The translational dynamics are modeled in a world-fixed inertia frame, given by:
\begin{align}
    \posdot &= \vel, \\
    \veldot &= \frac{1}{m}(T\zb + \faero + \fext) + \g,
\end{align}
where $\pos$ and $\vel$ are the position and velocity vectors; $T$ and $m$ are the collective thrust and total mass, respectively; $\g$ is the gravitational vector; $\faero$ indicates the aerodynamic drag force during flights, and $\fext$ lumps the external forces as disturbances; the vector $\zb \in \mathbb{R}^3$ is the unit vector along the body-frame z-axis. 

The rotational dynamics are expressed as:
\begin{align}
      \dot{\mathbf{R}} &= \mathbf{R} [\w]_\times, \\
    \wdot &= \invJ(-[\w]_ \times \J \w + \bm\tau + \tauext + \taua),
\end{align}
where $ \mathbf{R} \in \mathrm{SO}(3) $ is the rotation matrix from the body frame to the inertial frame, and $\w$ is the angular velocity expressed in the body frame. The operator $ [\cdot]_\times $ denotes the skew-symmetric matrix associated with the cross product. We denote $\J$ as the inertia matrix of the quadcopter; $\bm \tau$ is the torque generated by the rotors; $\taua$ and $\tauext$ are the aerodynamic torque and the disturbance torques, respectively. 

\begin{figure}
    \centering
    \includegraphics[width=0.5\linewidth]{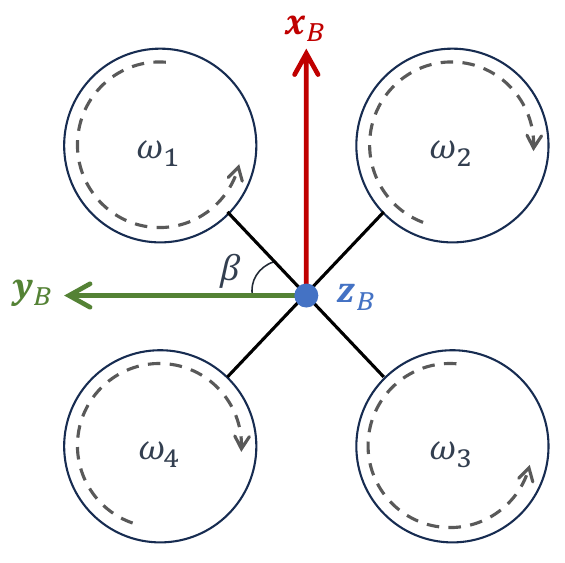}
    \caption{Illustration of the quadcopter's $x$-configuration motor layout in the body frame. The directions of rotation for each rotor ($\omega_1$ to $\omega_4$) and the geometric angle $\beta$ between the rotor arms and the body axis $\bm{x}_B$ are indicated.}
    \label{fig:fbd-drone}
\end{figure}

The collective thrust $T$ and torques $\bm \tau$ are functions of rotor speeds:
\begin{align}
\label{eq:thrusttorque2u}
\begin{bmatrix}
    T\\\bm \tau
\end{bmatrix} &= \bm G \bm u, \\
\label{eq:u2spd}
u_i &= \kt \, \omega_i^2, \quad i \in \{1, 2, 3, 4\},
\end{align}
where $ \mathbf{u} \in \mathbb{R}^4 $ is the vector of individual rotor thrusts, $ \omega_i $ is the speed of rotor $ i $, and $ k_t $ is the thrust coefficient.
$\bm G$ is a matrix defined as:
\begin{align} \label{eq:def-G}
    \bm G&= \begin{bmatrix}
        1 &1&1&1\\
        l \sin{\beta}&-l \sin{\beta}&-l \sin{\beta}&l \sin{\beta}\\
        -l \cos{\beta}&-l \cos{\beta}&l \cos{\beta}&l \cos{\beta}\\
        \ktau/\kt&-\ktau/\kt&\ktau/\kt&-\ktau/\kt
    \end{bmatrix},
\end{align}
where $l$ denotes the distance from the vehicle center to each rotor (arm length), and $\beta$ is the angle between the $x_B$ axis and the rotor arm direction, as illustrated in Figure~\ref{fig:fbd-drone}; $\ktau$ is the torque coefficient.

Quadcopter motors cannot instantaneously achieve commanded speeds due to rotor inertia and motor dynamics. Their transient behavior is commonly approximated using a first-order model, as adopted in prior work such as~\cite{eschmann2024datadrivenidentificationquadrotorssubject}:
\begin{align}
\dot{\msp} = \frac{1}{\tau_\mathrm{m}}(\msp_\mathrm{des}-\msp),    
\end{align}
where $\msp_\mathrm{des}$ is the desired motor speed; $\tau_\mathrm{m}$ is the motor time constant which can be identified using static
thrust stand testing.

The aerodynamic forces $\faero$ and torques $\taua$ together form the aerodynamic wrench, representing the collection of forces and moments generated by the quadcopter's motion through the air.
These effects become particularly significant in high-speed flights or strong wind conditions. While \textit{RotorPy} includes a parasite drag model identified on the Crazyflie, we adopt a planar drag formulation that has been validated across multiple quadcopter platforms and shown to transfer reliably to hardware~\cite{sun2024comparativestudynonlinearmpc, zhang2024learningbasedquadcoptercontrollerextreme}. This choice favors generalizability for benchmarking, while we still retain \textit{RotorPy}’s rotor drag and blade flapping terms.
The planar drag model is expressed as:
\begin{align}
    \mathbf{f}_{a,p}&=\begin{bmatrix}
        -\kdx v_{a,x}\\
        -\kdy v_{a,y}\\
        -\kdz v_{a,z} + \kh (v_{a,x}^2 + v_{a,y}^2)
    \end{bmatrix}, 
\end{align}
where $[v_{a,x},v_{a,y},v_{a,z}] = \mathbf{R}^T (\vel - \windvel)$ is the relative body airspeed with respect to the wind velocity $\windvel$; $k_{d,[x,y,z]}$ and $\kh$ are aerodynamic parameters configured within the simulator to reflect nominal drag and lift effects. 

In addition, we adopt the rotor drag and blade flapping models from \textit{RotorPy}, which account for aerodynamic effects due to rotor-induced flow and blade motion (see Eq. (8-9) in~\cite{folk2023rotorpy}).
\subsection{Control Library}

We provide a library of adaptive control methods to facilitate rapid prototyping and benchmarking for researchers.
Additionally, we include non-adaptive controllers as baselines to quantify performance improvements in the presence of external disturbances and model uncertainties.

The baseline controllers include: \begin{itemize} \item \texttt{geo}: a geometric controller~\cite{lee2011geo}; \item \texttt{mpc}: a nonlinear MPC~\cite{sun2024comparativestudynonlinearmpc}. \end{itemize}
The adaptive controllers include: \begin{itemize} \item \texttt{geo-a}: a geometric adaptive controller~\cite{goodarzi2015geoa}; \item \texttt{l1geo}: a geometric controller augmented with $\mathcal{L}_1$ adaptive control~\cite{wu2024l1quad}; \item \texttt{l1mpc}: a nonlinear MPC augmented with $\mathcal{L}_1$ adaptive control~\cite{taol1mpc}; \item \texttt{indi-a}: an incremental nonlinear dynamic inversion controller~\cite{tal2021indi}, which incorporates onboard adaptive parameter estimation to update control effectiveness~\cite{smeur2015adaptINDI}; \item \texttt{xadap}: a learning-based adaptive controller~\cite{zhang2024learningbasedquadcoptercontrollerextreme}. \end{itemize}

These controllers operate at different command levels: some, such as \texttt{geo-a} and \texttt{l1geo}, generate control commands at the collective thrust and moment level, while others, such as \texttt{xadap} and \texttt{indi-a}, directly output motor speed commands.
To standardize execution across different control levels, we compute the required motor speed for those not commanding it using~\eqref{eq:thrusttorque2u} and~\eqref{eq:u2spd}.
Note the model parameters such as $\kt$ and $\ktau$ used in these computations are not the ground-truth values but reference model parameters that may contain uncertainties. 
We discuss these uncertainties in detail in Section~\ref{sec:method-model-uncertainty}.

In addition, among these methods, \texttt{xadap} and \texttt{indi-a} are proposed as inner-loop controllers, which take intermediate control inputs such as commanded collective thrust and angular velocity from an outer-loop controller.
To complete the control pipeline, we concatenate it with a geometric controller as the outer-loop controller.

\subsection{Evaluation Task and Metrics}
We evaluate tracking performance along a desired trajectory as the primary task in our testbed.
For large-scale batch evaluations, we generate trajectories using randomized motion primitives following the method of~\cite{mueller2015trajectory}.
Specifically, each trajectory is composed of a sequence of short-duration polynomial segments, where the motion parameters such as velocity, acceleration, and segment duration are randomly sampled within dynamically feasible bounds.
This approach ensures physical feasibility, supports rapid generation, and produces a diverse set of trajectories that challenge different aspects of controller performance.
However, users can customize the testbed by specifying alternative trajectories tailored to their specific tasks, such as hovering or circular flight.
To quantify tracking accuracy, we use the root mean square error (RMSE) of position and the absolute error of heading as the primary precision metrics.
Additionally, we assess robustness using the success rate, where a flight is considered failed if its position $\pos$ violates the following spatial constraint at an
arbitrary instant:
\begin{align}
\label{eq:failure_cond}
    \|\pos_\mathrm{des}-\pos\|_2 \leq e_{\mathrm{x,max}}.
\end{align}
We select $e_{\mathrm{x,max}}=5$m in our testbed.
\subsection{External Disturbances}
To ensure a comprehensive evaluation, we implement a variety of disturbances commonly tested in previous studies such as~\cite{hanover2021performance,zhang2024learningbasedquadcoptercontrollerextreme,sun2024comparativestudynonlinearmpc}. 
\subsubsection{Gusty Wind}
Gusty wind varies rapidly, making it an effective test case for evaluating fast adaptation.
To model this effect, we use the Dryden wind turbulence model~\cite{etkin2005dynamics}, which characterizes turbulence based on the mean wind speed along each axis and its variance, defining the turbulence intensity.
The resulting disturbance acts as an aerodynamic wrench, as defined in~\ref{sec:quaddyna}, impacting both forces and moments on the quadcopter.
% \subsubsection{Off-center Payload}
% Off-center payloads are common disturbances in applications such as aerial logistics.
% %
% While in general such payloads shift the vehicle’s center of mass, which affects the dynamics beyond just added forces and torques, we do not explicitly model this effect.
% Instead, we approximate their impact as external forces and moments applied at the nominal center of mass, which is valid under near-hover conditions:
% \begin{align}
%     \fext &= m_\mathrm{p} \g,\\
%     \tauext &= \bm x_\mathrm{p} \times  \fext ,
% \end{align}
% where $m_\mathrm{p}$ and $\bm x_\mathrm{p}$ represent the payload mass and its relative location to the quadcopter's center of mass, respectively, both of which are randomized to introduce variability in the disturbance.
% The force and torque are applied simultaneously at a randomly selected time to simulate abrupt payload attachment or drop-off events.
\subsubsection{Off-center Payload}
Off-center payloads are common in aerial logistics and introduce coupled effects on mass, center of mass (COM), and inertia.
We model the payload as a rigidly attached point mass and account for its impact as follows:
\begin{enumerate} \item The total mass is updated:
\begin{align}
m_{\text{total}} &= m_{\text{vehicle}} + m_{\text{payload}},
\end{align}
 \item The center of mass (COM) is shifted to: 
 \begin{align}    
 \bm{x}_{\text{COM}} = \frac{m_{\text{vehicle}}\,\bm{x}_{\text{vehicle}} + m_{\text{payload}}\,\bm{x}_{\text{payload}}}{m_{\text{total}}},
 \end{align}
 where $\bm{x}_{\text{vehicle}}$ is the nominal COM and the origin of the body frame defined in Figure~\ref{fig:fbd-drone}. The shifted COM changes the rotor moment arms, so we also update the control allocation matrix $\bm{G}$ (defined in~\eqref{eq:def-G}) to reflect the new geometry.
 \item The inertia tensor is recomputed using the parallel axis theorem~\cite{meriam2015engineering}, accounting for the shifted mass and added point mass.
 \end{enumerate}

The gravitational force on the payload is naturally captured by the increased mass. However, the offset between the payload and the new COM introduces a torque due to gravity, computed as: \begin{align} \bm{\tau}_{\text{ext}} = [\bm{x}_{\text{payload}} - \bm{x}_{\text{COM}}]_ \times (m_{\text{payload}}\,\bm{g}).\end{align}

This torque is explicitly included as an external disturbance. To introduce variability, both the payload mass $m_{\text{payload}}$ and its attachment location $\bm{x}_{\text{payload}}$ are randomized across trials. To simulate pickup and drop-off events, the payload is attached or removed at a random time during flight.
\subsubsection{External Forces and Torques}
External force and torque disturbances are implemented as independent perturbations, applied separately (unlike inherently coupled forces and moments in the off-center payload case), allowing us to analyze their individual effects on the system.
Moreover, in contrast to the gusty wind case, these disturbances remain constant over time, enabling a controlled evaluation of their isolated impact without temporal variations.
\subsubsection{Loss of Rotor Effectiveness}
In real-world conditions, rotor effectiveness varies due to hardware differences, wear, or battery depletion.
To model this effect, we introduce a positive effectiveness factor $\bm{k_{\mathrm{eff}}}$, which scales the contribution of each motor and modifies the system dynamics as follows:
\begin{align}
\label{eq:moteff}
k_{\mathrm{eff},i} &= 1 + {\mathcal{U}}(-\delta,\delta), \\
    u_i &= k_{\mathrm{eff},i}\, \kt  \, \omega_i^2, \quad i \in \{1, 2, 3, 4\},
\end{align}
where $\mathcal{U}(-\delta,\delta)$ is a uniform distribution over $[-\delta,\delta]$; $\delta \in [0,1)$ defines the variation range of this motor's effectiveness. 

It is important to note that setting $k_{\mathrm{eff},i}$ to zero does not fully capture the dynamics of a complete rotor failure, as explored in~\cite{mueller2015rotorfailIJRR}.
In the event of motor failure, the vehicle loses yaw authority and experiences asymmetric drag. While these effects can be modeled with moderate extensions, such as including rotational drag terms, we do not implement them in this work.
\subsubsection{Latency} \label{sec:latency-def}
Latency refers to time delays between sensor measurements, controller decisions, and actuator execution. In quadcopters, such delays are typically due to communication overhead, processing constraints, or software pipeline structure. When present in feedback loops, latency can act as a form of disturbance that degrades control performance and may lead to instability.

To evaluate robustness to latency, we introduce a fixed control execution delay. Control inputs are buffered and applied after a specified delay, while sensor updates remain undelayed. This setup isolates the effect of actuation-side delay, resembling latency between a state estimator and controller or between ground station and vehicle.
The robustness of a controller to such delay can be characterized by its delay margin, defined as the maximum allowable latency before the system becomes unstable. While frequency-domain tools such as Bode or Nyquist plots can estimate delay margins for linear systems~\cite{franklin2015feedback}, they are less effective for nonlinear systems like quadcopters.

Instead, we define a delay margin as the maximum control delay under which the controller can complete a trajectory with the position error contained within a user-defined threshold at any time:
\begin{align}
\|\pos_{\mathrm{des}}-\pos\|_2  \leq \ethresh, \end{align} where $\ethresh$ is a user-defined position error threshold. This value can be tuned depending on the scale and precision requirements of the application; for example, smaller thresholds (e.g., 0.5m) for micro aerial vehicles and larger ones (e.g., 2–5m) for full-scale platforms. We estimate this margin by incrementally increasing the control delay and recording the point at which the system violates the threshold across different trajectory types.
\subsection{Model Uncertainties} \label{sec:method-model-uncertainty}
Beyond external disturbances, a quadcopter’s performance is influenced by uncertainties in the reference model, such as parameter offsets or errors. We consider two types of model uncertainty: \textit{Uniform} and \textit{Scaled}.

\subsubsection{Uniform Uncertainties}
In this type of uncertainties, each parameter is perturbed independently around its nominal value using a uniform distribution within a fixed range, which models unbiased parameter errors.

\subsubsection{Scaled Uncertainties} \label{sec:model-uncertainty-scaled}
When seeking a significantly different reference model, uniform perturbations may result in physically inconsistent configurations. For example, sampling mass and inertia independently from a uniform distribution over $\pm$ 50\% may yield combinations where the mass is reduced while the inertia is increased, violating physical relationships between size, mass distribution, and dynamics. To address this issue, we introduce a scaling factor $c$ that proportionally adjusts the arm length:
\begin{align}
c &= \mathcal{U}(-c_{\max}, c_{\max}), \, c_{\max}\in[0,1)\\
    l_{\mathrm{ref}}&=(1+c)\,l_{\mathrm{true}},
\end{align}
where $l_{\mathrm{ref}}$ is the arm length of the reference model; $l_{\mathrm{true}}$ is the ground truth value of the arm length. The range of the scaling factor $c$ guarantees that the resulting arm length remains strictly positive.

Assuming constant density and isotropic scaling, we scale mass with the cube of the arm length, inertia with the fifth power, and drag with the square. To reflect the relationship between quadcopter size and motor strength, we choose to exponentially scale the thrust coefficient with $c$, ensuring that larger quadcopters have proportionally stronger motors. Other parameters, including maximum motor speed and propeller constants, are scaled linearly. 
This structured scaling preserves physical consistency, preventing unrealistic configurations such as small drones with excessively powerful motors. The single scaling factor $c$ quantifies the deviation of the reference model from the ground truth, with $\max|c|$ representing the maximum allowable deviation.

Finally, to account for deviations from idealized scaling laws, we introduce uniformly distributed noise in the range [-20\%, 20\%] across all parameters after applying the scaling factor $c$. This enhances model diversity while maintaining plausibility.

\section{Evaluation Examples} \label{sec:eva}
In this section, we present examples of evaluations conducted using our testbed, which supports automated generation of statistical tables and plots. To demonstrate the framework’s versatility, we show a subset of controllers for each disturbance type. This selection highlights the range of controllers supported by the testbed while ensuring that each example remains focused and interpretable.
For all results presented in this section, we adopt the \textit{large quadrotor} model from~\cite{zhang2024learningbasedquadcoptercontrollerextreme}, using the parameters specified in Table~\ref{tab:quad-params}.
\begin{table}
\centering
\caption{Quadrotor Parameters}
\begin{tabularx}{\linewidth}{cc|c}
        \toprule
         \multicolumn{2}{c|}{Parameter}  & Value
      \\
        \midrule
        $m$ &[kg]& $0.75$ \\
        $l$ &[m]&$0.166$\\
        $\beta$&[deg]&$45$\\
        $\bm J$ &[kg$\cdot$m$^2$] &diag($0.0047,0.005,0.0074$)\\
        $(k_{d,x},k_{d,y},k_{d,z})$&[kg/s]&$(0.62,0.62,0.62)$\\
        $\kh$&[kg/m]&0.0\\
        $\kt$&[N/(rad/s)$^2$]&$7.64e^{-6}$\\
        $\ktau$&[N$\cdot$m/(rad/s)$^2$]&$1.07e^{-7}$\\
        $\tau_{\mathrm{m}}$&[s]&$0.01$\\
        $\omega_{\max}$&[rad/s]&1000\\
       
        \bottomrule
    \end{tabularx}
    \label{tab:quad-params}
\end{table} 
%
% Table II: Experiment Type vs Randomization Range

\subsection{Uncertainty-free Condition}
First, we present the performance of each controller in the absence of disturbances and model uncertainties.
We fly the vehicle using a reference model that best represents our knowledge of the system dynamics.
In this case, model parameters such as mass, inertia, and rotor arm lengths are assumed to be perfectly known, which means that the parameters used in model-based controllers match exactly with those in the simulation.
These experiments help establish the best possible performance each method can achieve under the uncertainty/disturbance-free conditions.
The only exception is the learning-based controller \texttt{xadap}, which does not rely on a reference model due to its design.

\lstset{style=terminal}
\begin{lstlisting}
$ python run_eval.py --controller all --experiment no
\end{lstlisting}

This command runs the evaluation task using all available controllers under this uncertainty-free conditions and reports task statistics, such as position RMSE, heading error, and success rate, based on the metrics defined in the previous section.
Table~\ref{tab:ideal-benchmark} shows baseline performance under idealized conditions. All controllers, except the learning-based \texttt{xadap}, depend on parameter tuning and can achieve comparable tracking accuracy and success rates in the absence of disturbances and model uncertainties.
\begin{table}
\centering
\caption{Performance of each controller in the absence of disturbances and model uncertainties. With appropriate tuning, all controllers can achieve similar performance.
}
\begin{tabularx}{\linewidth}{l|ccc}
        \toprule
        Method & Position RMSE & Heading Error & Success Rate \\ 
        & [m]& [deg]&\\
        \midrule
        \texttt{geo}   & $0.074 \pm 0.016$ & $0.523 \pm 0.137$ & 100\% \\ 
        \texttt{geo-a} & $0.065 \pm 0.016$ & $0.565 \pm 0.148$ & 100\% \\ 
        \texttt{l1geo} & $0.048 \pm 0.015$ & $0.223 \pm 0.156$ & 100\% \\ 
        \texttt{indi-a} & $0.030 \pm 0.007$ & $0.254 \pm 0.157$ & 100\% \\ 
        \texttt{l1mpc} & $0.035 \pm 0.012$ & $0.224 \pm 0.123$ & 100\% \\ 
        \texttt{mpc}   & $0.040 \pm 0.012$ & $0.890 \pm 0.285$ & 100\% \\ 
        \texttt{xadap} & $0.046 \pm 0.009$ & $14.044 \pm 0.648$ & 100\% \\ 
        \bottomrule
    \end{tabularx}
    \label{tab:ideal-benchmark}
\end{table}

\subsection{Gusty Wind}
\begin{figure}
    \centering
    \includegraphics[width=\linewidth]{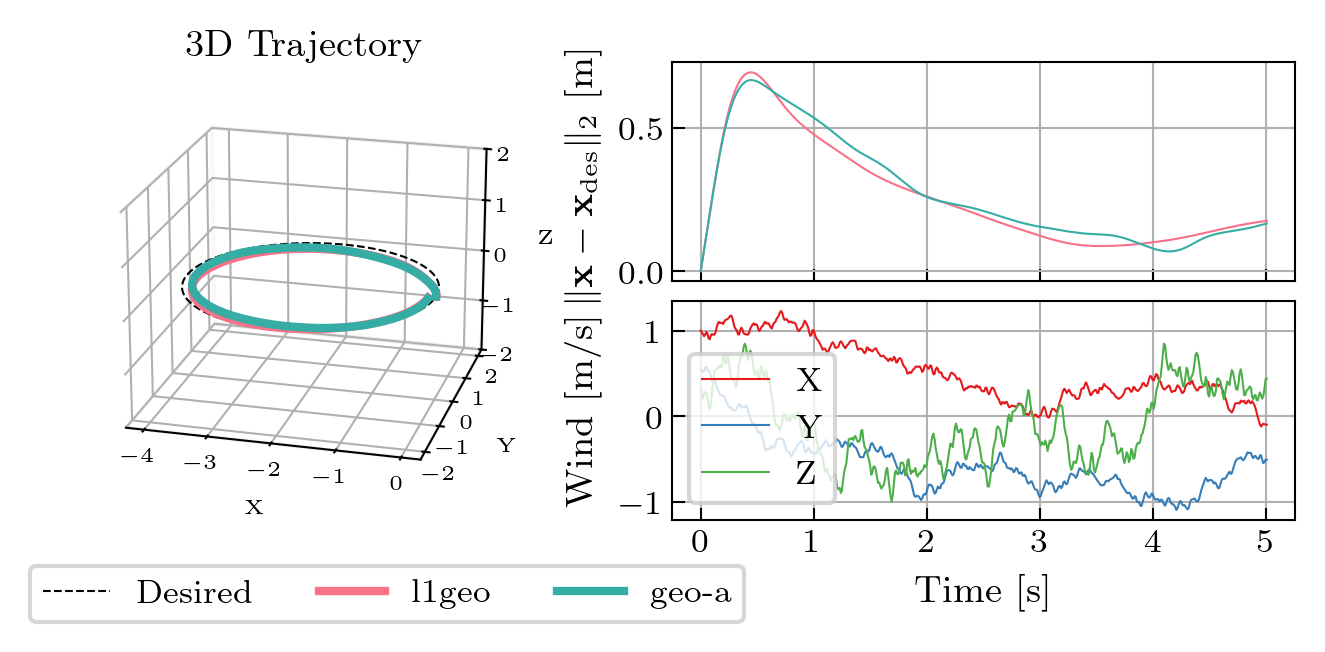}
    \caption{Position tracking of \texttt{l1geo} and \texttt{geo-a} along a circular trajectory in the presence of gusty wind, generated using the Dryden turbulence model.
    % \textbf{Top}: Euclidean norm of position error, \textbf{Middle}: wind velocity, \textbf{bottom}: average motor speed over 4 motors. 
    }
    \label{fig:wind-example}
\end{figure}
We now demonstrate how the framework can be used to evaluate controller performance under wind disturbances.  
In this example, we introduce a gusty wind field modeled using the Dryden turbulence model, which varies continuously in time and direction.  
Rather than comparing methods via aggregated tables, we focus on visualizing performance along multiple dimensions, such as position error, control effort, and system response, on a single trajectory.  
To visualize the performance of selected methods on a single rollout, we can enable the \texttt{--vis} flag in the evaluation script:  
\lstset{style=terminal}
\begin{lstlisting}
$ python run_eval.py --controller l1geo geo-a --experiment wind --trajectory circle --vis
\end{lstlisting}
This example, along with Figure~\ref{fig:wind-example} generated, shows the position tracking performance of the \texttt{l1geo} and \texttt{geo-a} controllers along a circular trajectory under gusty wind conditions.
Similarly, external force and torque disturbances can be evaluated by modifying the command line argument accordingly, where the disturbance plot will update to reflect the corresponding disturbance type.  

\subsection{Off-center Payload}
\begin{figure}
    \centering
    \includegraphics[width=\linewidth]{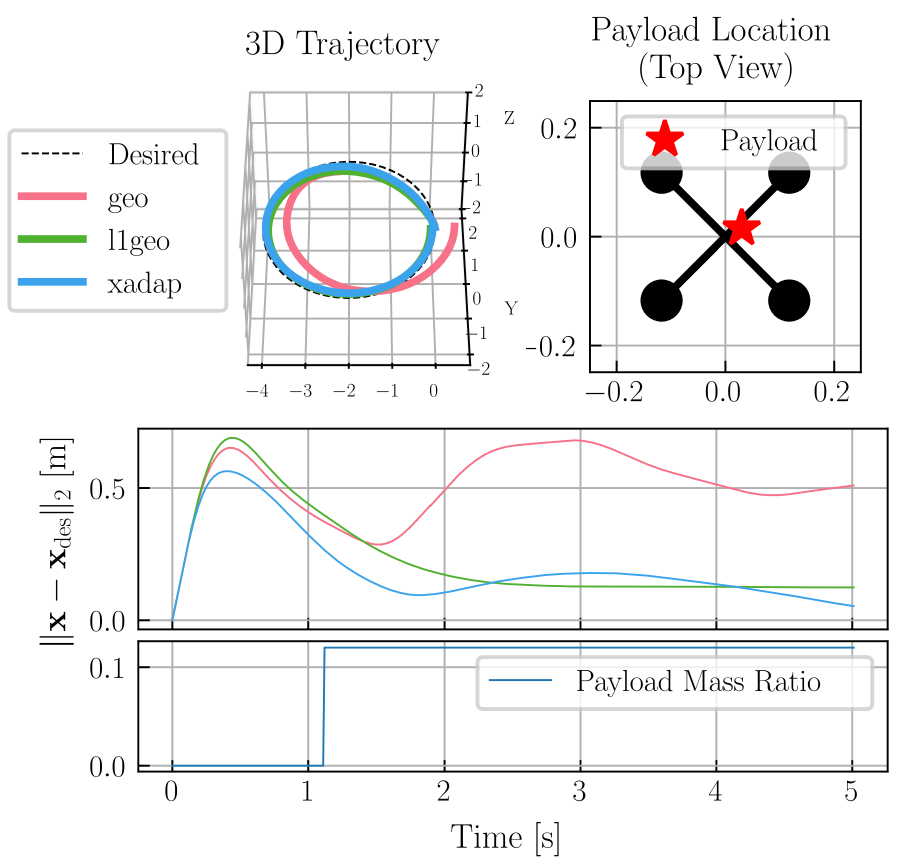}
    \caption{Position tracking of \texttt{geo}, \texttt{l1geo}, and \texttt{xadap} under a circular trajectory with an off-center payload randomly attached mid-flight. The payload introduces coupled effects on center-of-mass, inertia, and external torque. The deviation from the desired trajectory before payload attachment arises from the transient controller response as the vehicle transitions from hovering to trajectory tracking.}
    \label{fig:payload-example}
\end{figure}
\lstset{style=terminal}
\begin{lstlisting}
$ python run_eval.py --controller geo l1geo xadap --vis --experiment payload --trajectory circle
\end{lstlisting}

This command generates Figure~\ref{fig:payload-example}, which visualizes the position tracking performance of \texttt{geo}, \texttt{l1geo}, and \texttt{indi-a}.  
The figure also includes the payload's location and the ratio of the payload mass to the vehicle's mass over time.

\subsection{Loss of Rotor Effectiveness}
\begin{figure}
    \centering
    \includegraphics[width=\linewidth]{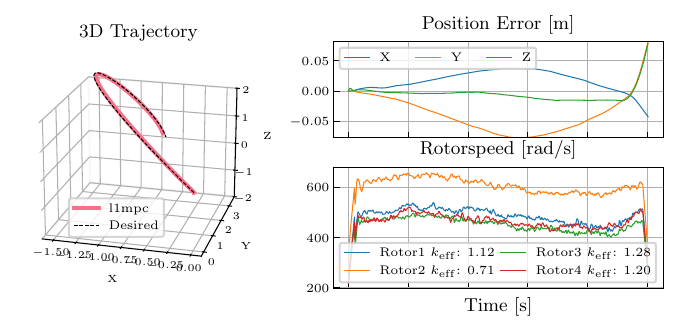}
    \caption{Position tracking and motor speed response of the \texttt{l1mpc} controller on a randomized trajectory under rotor effectiveness variations. Each rotor’s effectiveness factor is perturbed, simulating actuator-level faults.}
    \label{fig:rotoreff-example}
\end{figure}

Figure~\ref{fig:rotoreff-example} illustrates the motor speed response and tracking performance of the \texttt{l1mpc} controller when following a randomized motion primitive trajectory under rotor effectiveness variations.  
Additionally, the perturbed effectiveness coefficient for each rotor is plotted, providing insight into how motor effectiveness influences system behavior.  
This visualization is generated by running:  

\lstset{style=terminal}
\begin{lstlisting}
$ python run_eval.py --controller l1mpc --vis --experiment rotoreff
\end{lstlisting}

\subsection{Latency}
\lstset{style=terminal} 
\begin{lstlisting}
$ python run_eval.py --controller geo --delay_margin 
\end{lstlisting}

\begin{figure} \centering \includegraphics[width=\linewidth]{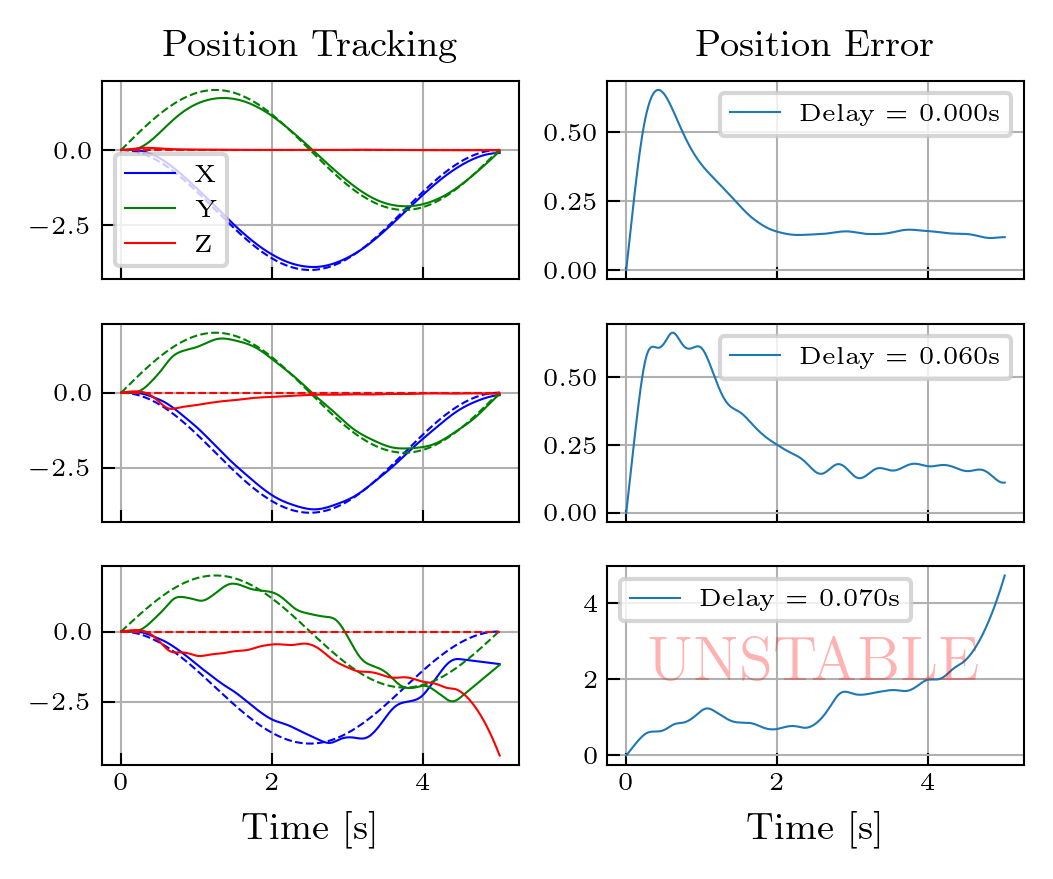} \caption{Position tracking of the \texttt{geo} controller on a circular trajectory under increasing control latency. Three cases are shown: no delay, the maximum stable delay, and the first unstable delay instance.} \label{fig:latency-example} \end{figure}

This command generates Figure~\ref{fig:latency-example}, which illustrates how control latency affects tracking performance under a circular trajectory using the \texttt{geo} controller. The plots show position tracking and position error under three latency settings: 0.00s (no delay), 0.06s (the final delay before instability), and 0.07s (the smallest delay at which instability occurs). 
The observed failure point is specific to the chosen controller, trajectory, and error threshold. Different controllers or trajectories may tolerate different amounts of latency before instability emerges. In addition, the delay margin is defined relative to a user-specified error threshold (Section~\ref{sec:latency-def}); varying this threshold will also affect the measured margin.

\subsection{Model Uncertainty}
\begin{figure}
    \centering
    \includegraphics[width=\linewidth]{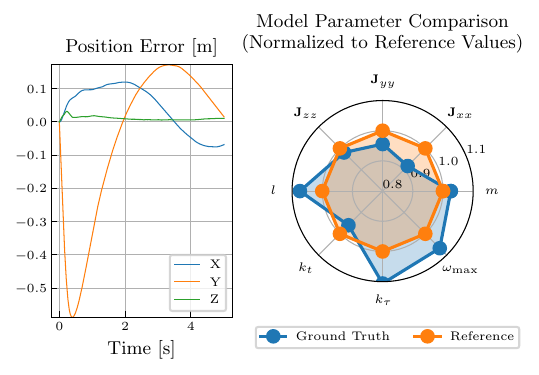}
    \caption{Left: Position tracking of \texttt{xadap} under structured model parameter perturbations.
Right: Polar chart comparing reference model parameters to ground truth, normalized to nominal values.}
    \label{fig:model-uncertainty}
\end{figure}

\lstset{style=terminal}
\begin{lstlisting}
$ python run_eval.py --controller xadap --vis --experiment uncertainty
\end{lstlisting}

This example (Figure~\ref{fig:model-uncertainty}) visualizes both the tracking performance of \texttt{xadap} and the impact of model uncertainties. 
The model uncertainties are visualized by a polar chart illustrating the discrepancy between the reference model parameters used by the controller and the actual ground-truth parameters, providing insight into the magnitude of modeling errors across key parameters.

\section{Extreme Adaptation Evaluation}  
The previous section demonstrated how our testbed can simulate a variety of disturbance types and model uncertainties, enabling qualitative inspection of controller behavior under each condition.
In this section, we shift focus toward evaluating the limits of controller robustness by progressively increasing the intensity of a disturbance to identify failure points. 
This type of evaluation has been explored in previous works~\cite{zhang2024learningbasedquadcoptercontrollerextreme,hanover2021performance,antonio2021wildhighspd}, where the disturbance magnitude or uncertainty range is systematically increased (e.g., increasing the maximum trajectory speed or expanding the model uncertainty bounds) to determine the threshold at which performance degrades below an acceptable level.  
Such evaluations serve as valuable tools for assessing robustness and improving controller design.  
Additionally, they are particularly useful for analyzing the out-of-distribution performance of learning-based methods.  

To facilitate this analysis, we implement an automatic stress-testing feature in our testbed.  
This feature progressively increases the range of randomization until either the success rate of the selected method drops to a selected threshold such as 10\% or the maximum disturbance intensity is reached.  
The type of disturbance increment depends on the specific experiment setting:  

\begin{itemize}  
    \item \textbf{Wind}: Increase in gust magnitude  
    \item \textbf{Payload}: Increase in mass (but not arm length, as arbitrarily large offsets are unrealistic)  
    \item \textbf{External Force}: Increase in force magnitude  
    \item \textbf{External Torque}: Increase in torque magnitude  
    \item \textbf{Loss of Rotor Effectiveness}: Increase in the maximum deviation of the motor effectiveness vector from a nominal, identical-motor setup, as quantified by $\delta$ in~\eqref{eq:moteff}.
    \item \textbf{Model Uncertainty}: Increase in the maximum deviation from the ground-truth model, quantified by $\max|c|$ in Section~\ref{sec:model-uncertainty-scaled}.  
\end{itemize}  
\begin{figure}
    \centering
    \includegraphics[width=\linewidth]{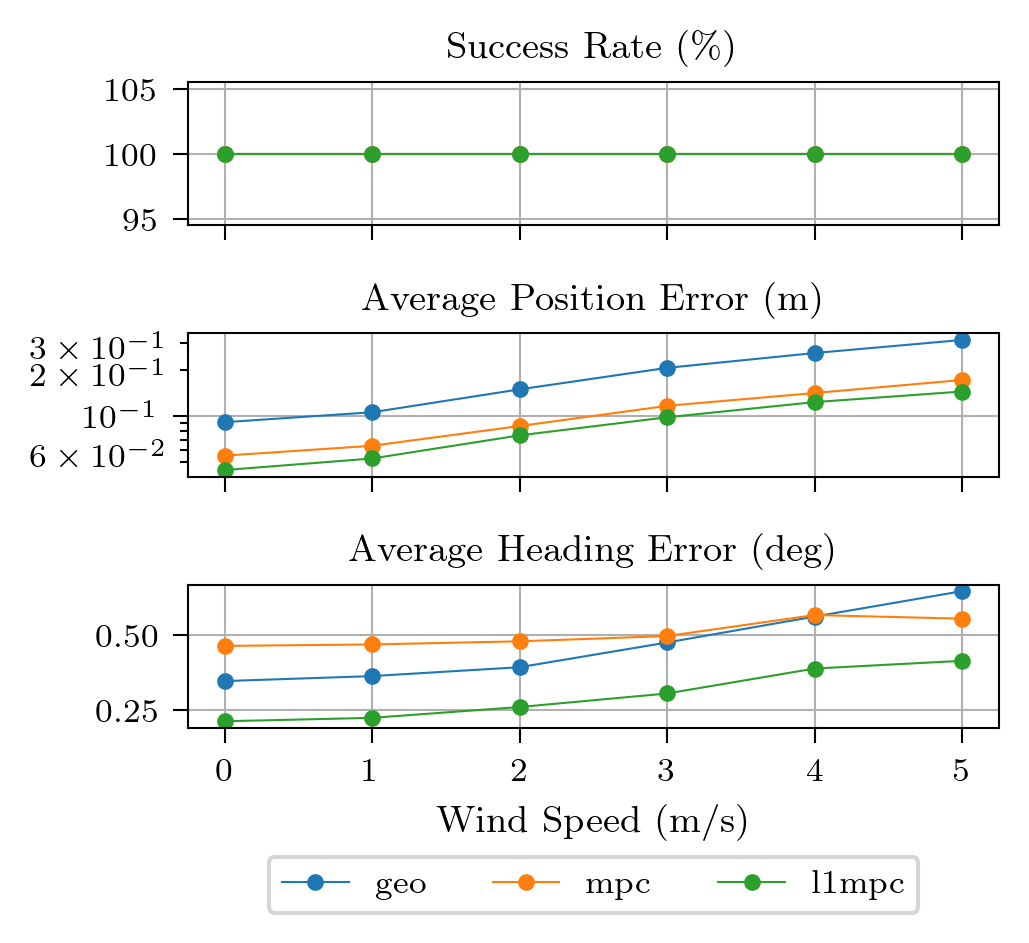}
    \caption{Tracking error and heading error as a function of wind speed for \texttt{geo}, \texttt{mpc}, and \texttt{l1mpc}, under increasing gust intensity using automated stress testing.}
    \label{fig:when2fail-wind}
\end{figure}
For example, we can evaluate the performance of \texttt{geo}, \texttt{mpc}, and \texttt{l1mpc} over 100 trials with gusty wind speed increasing from 0 to 5 m/s in increments of 1 m/s using the following command:
\begin{lstlisting}
$ python run_eval.py --controller geo mpc l1mpc --experiment wind --when2fail --max_intensity 5 --intensity_step 1
\end{lstlisting}
The results, shown in Figure~\ref{fig:when2fail-wind}, indicate that while all methods maintain a 100\% success rate, position tracking error and heading error increase as wind speed intensifies.

\begin{figure}
    \centering
    \includegraphics[width=\linewidth]{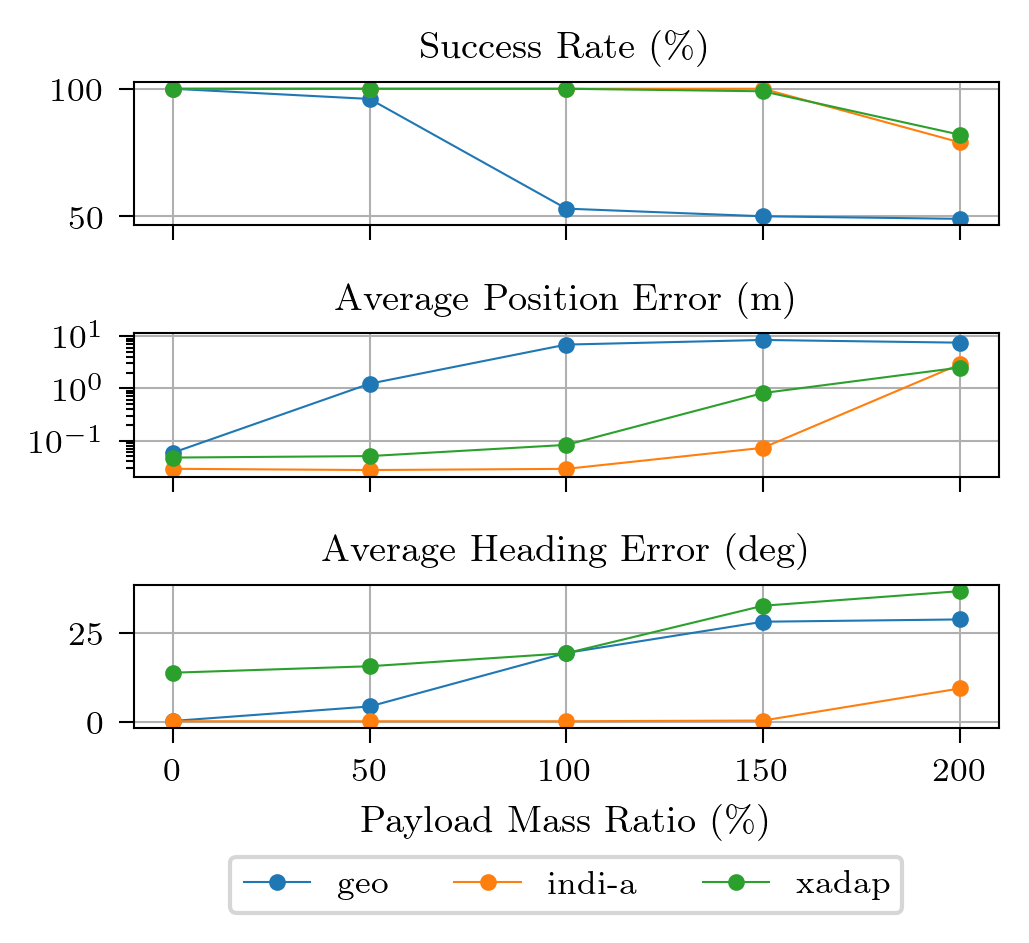}
    \caption{Tracking performance of \texttt{geo}, \texttt{indi-a}, and \texttt{xadap} under increasing payload mass ratios using automated stress testing.}
    \label{fig:when2fail-payload}
\end{figure}

Another example (Figure~\ref{fig:when2fail-payload}) examines how increasing the payload mass ratio impacts controller performance. The following command runs the evaluation:

\lstset{style=terminal} \begin{lstlisting} 
$ python run_eval.py --controller geo indi-a xadap --experiment payload --when2fail --max_intensity 2 --intensity_step 0.5 \end{lstlisting}

This command increases the maximum possible payload mass ratio (relative to the nominal vehicle mass) from 0\% to 200\% in increments of 50\%. The results show that while the non-adaptive \texttt{geo} controller degrades significantly as payload mass increases, both \texttt{indi-a} and \texttt{xadap} maintain high success rates and better tracking performance under high payload disturbances.
\section{Conclusion}
This paper introduces a simulation framework for benchmarking quadcopter controllers under external disturbances and dynamic uncertainties, including wind, payload, rotor faults, and latency. Built on \textit{RotorPy}, the framework enables reproducible testing across different robust adaptive controllers and provides a set of evaluation tools for performance and robustness analysis. It supports repeatable and extensible experimentation, contributing to ongoing efforts in reproducible robotics research.  

The framework is designed to be extensible with additional controllers and disturbance types. Future work may include incorporating more diverse wind models, simulating complete motor failures, and supporting additional real-world uncertainties. These extensions would further enhance the framework’s ability to evaluate adaptive controllers under challenging and realistic conditions.
\section{Acknowledgement}
This work was supported by the Hong Kong Center for Logistics Robotics, NASA Cooperative Agreement (80NSSC20M0229), NASA ULI (80NSSC22M0070), and AFOSR (FA955021-1-0411).
The experimental testbed at the HiPeRLab is the result of contributions of many people, a full list of which can be found at \url{hiperlab.berkeley.edu/members/}.

\balance
\bibliographystyle{IEEEtran}
\bibliography{ref}

\begin{thebibliography}{10}
\providecommand{\url}[1]{#1}
\csname url@samestyle\endcsname
\providecommand{\newblock}{\relax}
\providecommand{\bibinfo}[2]{#2}
\providecommand{\BIBentrySTDinterwordspacing}{\spaceskip=0pt\relax}
\providecommand{\BIBentryALTinterwordstretchfactor}{4}
\providecommand{\BIBentryALTinterwordspacing}{\spaceskip=\fontdimen2\font plus
\BIBentryALTinterwordstretchfactor\fontdimen3\font minus \fontdimen4\font\relax}
\providecommand{\BIBforeignlanguage}[2]{{%
\expandafter\ifx\csname l@#1\endcsname\relax
\typeout{** WARNING: IEEEtran.bst: No hyphenation pattern has been}%
\typeout{** loaded for the language `#1'. Using the pattern for}%
\typeout{** the default language instead.}%
\else
\language=\csname l@#1\endcsname
\fi
#2}}
\providecommand{\BIBdecl}{\relax}
\BIBdecl

\bibitem{SAVIOLO202345learningagilereview}
A.~Saviolo and G.~Loianno, ``Learning quadrotor dynamics for precise, safe, and agile flight control,'' \emph{Annual Reviews in Control}, vol.~55, pp. 45--60, 2023.

\bibitem{johnson2005pid}
M.~A. Johnson and M.~H. Moradi, \emph{PID control}.\hskip 1em plus 0.5em minus 0.4em\relax Springer, 2005.

\bibitem{lee2011geo}
T.~Lee, M.~Leok, and N.~Mcclamroch, ``Geometric tracking control of a quadrotor uav on se(3),'' 01 2011, pp. 5420 -- 5425.

\bibitem{camacho2013mpc-book}
E.~F. Camacho and C.~B. Alba, \emph{Model predictive control}.\hskip 1em plus 0.5em minus 0.4em\relax Springer Science \& Business Media, 2013.

\bibitem{rawlings2020mpc-book}
J.~B. Rawlings, D.~Q. Mayne, and M.~M. Diehl, \emph{{Model Predictive Control: Theory, Computation, and Design, 2nd Ed.}}\hskip 1em plus 0.5em minus 0.4em\relax Nob Hill Publishing, 2020.

\bibitem{hovakimyan2010l1}
N.~Hovakimyan and C.~Cao, \emph{L1 adaptive control theory: Guaranteed robustness with fast adaptation}.\hskip 1em plus 0.5em minus 0.4em\relax SIAM, 2010.

\bibitem{wu2024l1quad}
Z.~Wu, S.~Cheng, P.~Zhao, A.~Gahlawat, K.~A. Ackerman, A.~Lakshmanan, C.~Yang, J.~Yu, and N.~Hovakimyan, ``L1quad: L1 adaptive augmentation of geometric control for agile quadrotors with performance guarantees,'' \emph{IEEE Transactions on Control Systems Technology}, vol.~33, no.~2, pp. 597--612, 2025.

\bibitem{goodarzi2015geoa}
F.~A. Goodarzi, D.~Lee, and T.~Lee, ``Geometric adaptive tracking control of a quadrotor unmanned aerial vehicle on se(3) for agile maneuvers,'' \emph{Journal of Dynamic Systems, Measurement, and Control}, vol. 137, no.~9, Jun. 2015.

\bibitem{smeur2015adaptINDI}
E.~Smeur, Q.~Chu, and G.~Croon, ``Adaptive incremental nonlinear dynamic inversion for attitude control of micro air vehicles,'' \emph{Journal of Guidance, Control, and Dynamics}, vol.~39, pp. 1--12, 12 2015.

\bibitem{zhang2024learningbasedquadcoptercontrollerextreme}
\BIBentryALTinterwordspacing
D.~Zhang, A.~Loquercio, J.~Tang, T.-H. Wang, J.~Malik, and M.~W. Mueller, ``A learning-based quadcopter controller with extreme adaptation,'' \emph{IEEE Transactions on Robotics}, vol.~41, p. 3948–3964, 2025. [Online]. Available: \url{http://dx.doi.org/10.1109/TRO.2025.3577037}
\BIBentrySTDinterwordspacing

\bibitem{kaufmann2022benchmark}
E.~Kaufmann, L.~Bauersfeld, and D.~Scaramuzza, ``A benchmark comparison of learned control policies for agile quadrotor flight,'' in \emph{2022 International Conference on Robotics and Automation (ICRA)}, 2022, pp. 10\,504--10\,510.

\bibitem{Bauersfeld__2021}
L.~Bauersfeld*, E.~Kaufmann*, P.~Foehn, S.~Sun, and D.~Scaramuzza, ``Neurobem: Hybrid aerodynamic quadrotor model,'' in \emph{Robotics: Science and Systems XVII}, ser. RSS2021.\hskip 1em plus 0.5em minus 0.4em\relax Robotics: Science and Systems Foundation, Jul. 2021.

\bibitem{zhang2024proxflyrobustcontrolclose}
\BIBentryALTinterwordspacing
R.~Zhang, D.~Zhang, and M.~W. Mueller, ``Proxfly: Robust control for close proximity quadcopter flight via residual reinforcement learning,'' 2024. [Online]. Available: \url{https://arxiv.org/abs/2409.13193}
\BIBentrySTDinterwordspacing

\bibitem{hanover2021performance}
D.~Hanover, P.~Foehn, S.~Sun, E.~Kaufmann, and D.~Scaramuzza, ``Performance, precision, and payloads: Adaptive nonlinear mpc for quadrotors,'' \emph{IEEE Robotics and Automation Letters}, vol.~7, no.~2, pp. 690--697, 2022.

\bibitem{Furrer2016rotorS}
F.~Furrer, M.~Burri, M.~Achtelik, and R.~Siegwart, \emph{RotorS---A Modular Gazebo MAV Simulator Framework}.\hskip 1em plus 0.5em minus 0.4em\relax Cham: Springer International Publishing, 2016, pp. 595--625.

\bibitem{sun2024comparativestudynonlinearmpc}
\BIBentryALTinterwordspacing
S.~Sun, A.~Romero, P.~Foehn, E.~Kaufmann, and D.~Scaramuzza, ``A comparative study of nonlinear mpc and differential-flatness-based control for quadrotor agile flight,'' 2024. [Online]. Available: \url{https://arxiv.org/abs/2109.01365}
\BIBentrySTDinterwordspacing

\bibitem{Foehn22scienceagilicious}
P.~Foehn, E.~Kaufmann, A.~Romero, R.~Penicka, S.~Sun, L.~Bauersfeld, T.~Laengle, G.~Cioffi, Y.~Song, A.~Loquercio, and D.~Scaramuzza, ``Agilicious: Open-source and open-hardware agile quadrotor for vision-based flight,'' \emph{AAAS Science Robotics}, 2022.

\bibitem{song2020flightmare}
Y.~Song, S.~Naji, E.~Kaufmann, A.~Loquercio, and D.~Scaramuzza, ``Flightmare: A flexible quadrotor simulator,'' in \emph{Conference on Robot Learning}, 2020.

\bibitem{panerati2021learningtofly}
J.~Panerati, H.~Zheng, S.~Zhou, J.~Xu, A.~Prorok, and A.~P. Schoellig, ``Learning to fly---a gym environment with pybullet physics for reinforcement learning of multi-agent quadcopter control,'' in \emph{2021 IEEE/RSJ International Conference on Intelligent Robots and Systems (IROS)}, 2021, pp. 7512--7519.

\bibitem{zha2024agrifly}
J.~Zha, T.~Yang, and M.~W. Mueller, ``Agri-fly: Simulator for uncrewed aerial vehicle flight in agricultural environments,'' \emph{IEEE Access}, vol.~12, pp. 140\,900--140\,907, 2024.

\bibitem{folk2023rotorpy}
S.~Folk, J.~Paulos, and V.~Kumar, ``{RotorPy}: A python-based multirotor simulator with aerodynamics for education and research,'' \emph{arXiv preprint arXiv:2306.04485}, 2023.

\bibitem{yuan2022safecontrolgymunifiedbenchmarksuite}
\BIBentryALTinterwordspacing
Z.~Yuan, A.~W. Hall, S.~Zhou, L.~Brunke, M.~Greeff, J.~Panerati, and A.~P. Schoellig, ``safe-control-gym: a unified benchmark suite for safe learning-based control and reinforcement learning in robotics,'' 2022. [Online]. Available: \url{https://arxiv.org/abs/2109.06325}
\BIBentrySTDinterwordspacing

\bibitem{Mueller2025Dynamics}
\BIBentryALTinterwordspacing
M.~W. Mueller, \emph{Dynamics and Control of Autonomous Flight}, 1st~ed., ser. Mathematical Engineering.\hskip 1em plus 0.5em minus 0.4em\relax Springer Cham, 2025. [Online]. Available: \url{https://doi.org/10.1007/978-3-031-91871-1}
\BIBentrySTDinterwordspacing

\bibitem{eschmann2024datadrivenidentificationquadrotorssubject}
\BIBentryALTinterwordspacing
J.~Eschmann, D.~Albani, and G.~Loianno, ``Data-driven system identification of quadrotors subject to motor delays,'' 2024. [Online]. Available: \url{https://arxiv.org/abs/2404.07837}
\BIBentrySTDinterwordspacing

\bibitem{taol1mpc}
R.~Tao, P.~Zhao, I.~Kolmanovsky, and N.~Hovakimyan, ``Robust adaptive mpc using uncertainty compensation,'' in \emph{2024 American Control Conference (ACC)}, 2024, pp. 1873--1878.

\bibitem{tal2021indi}
E.~Tal and S.~Karaman, ``Accurate tracking of aggressive quadrotor trajectories using incremental nonlinear dynamic inversion and differential flatness,'' \emph{IEEE Transactions on Control Systems Technology}, vol.~29, no.~3, pp. 1203--1218, 2021.

\bibitem{mueller2015trajectory}
M.~W. Mueller, M.~Hehn, and R.~D'Andrea, ``A computationally efficient motion primitive for quadrocopter trajectory generation,'' \emph{IEEE Transactions on Robotics}, vol.~31, no.~6, pp. 1294--1310, 2015.

\bibitem{etkin2005dynamics}
B.~Etkin, \emph{Dynamics of Atmospheric Flight}.\hskip 1em plus 0.5em minus 0.4em\relax Dover Publications, 2005.

\bibitem{meriam2015engineering}
J.~L. Meriam and L.~G. Kraige, \emph{Engineering Mechanics: Dynamics}, 8th~ed.\hskip 1em plus 0.5em minus 0.4em\relax Hoboken, NJ: Wiley, 2015.

\bibitem{mueller2015rotorfailIJRR}
M.~W. Mueller and R.~D'Andrea, ``Relaxed hover solutions for multicopters: Application to algorithmic redundancy and novel vehicles,'' \emph{The International Journal of Robotics Research}, vol.~35, no.~8, pp. 873--889, 2016.

\bibitem{franklin2015feedback}
G.~F. Franklin, J.~D. Powell, and A.~Emami-Naeini, \emph{Feedback Control of Dynamic Systems}, 7th~ed.\hskip 1em plus 0.5em minus 0.4em\relax Pearson, 2015.

\bibitem{antonio2021wildhighspd}
A.~Loquercio, E.~Kaufmann, R.~Ranftl, M.~Müller, V.~Koltun, and D.~Scaramuzza, ``Learning high-speed flight in the wild,'' \emph{Science Robotics}, vol.~6, no.~59, p. eabg5810, 2021.

\end{thebibliography}

\end{document}